\documentclass{article}

\usepackage[square, numbers]{natbib}
    \PassOptionsToPackage{numbers, compress}{natbib}
\usepackage{graphicx}
\usepackage{multirow}
\usepackage[table,xcdraw,dvipsnames]{xcolor}

\usepackage[preprint]{neurips_2020}

\usepackage[utf8]{inputenc} % allow utf-8 input
\usepackage[T1]{fontenc}    % use 8-bit T1 fonts
\usepackage{hyperref}       % hyperlinks
\usepackage{url}            % simple URL typesetting
\usepackage{booktabs}       % professional-quality tables
\usepackage{amsfonts}       % blackboard math symbols
\usepackage{nicefrac}       % compact symbols for 1/2, etc.
\usepackage{microtype}      % microtypography
\usepackage{xcolor}    
\usepackage{tabularx}% colors

\usepackage{hyperref}
\usepackage{url}

\usepackage{enumitem} 
\usepackage{tabularx}
\usepackage{booktabs}
\usepackage{multirow} 
\usepackage{color}

\usepackage{epsfig}
\usepackage{amsmath}
\usepackage{amssymb}
\usepackage{algorithm,algpseudocode}
\usepackage{tabularx}
\usepackage{booktabs}
\usepackage{makecell}
\usepackage{tabularx}
\usepackage{listings}

\usepackage{graphicx}
\usepackage{balance}
\usepackage{adjustbox}
\usepackage{mathtools}
\usepackage{color}
\usepackage{wrapfig}

\usepackage{epsfig}
\usepackage{amsmath}
\usepackage{amssymb}
\usepackage{algorithm,algpseudocode}
\usepackage{tabularx}
\usepackage{booktabs}
\usepackage{makecell}
\usepackage{tabularx}
\usepackage{listings}
\definecolor{pcolor}{HTML}{F26522}
\definecolor{ocolor}{HTML}{50B7C1}

\newcommand{\tablestyle}[2]{\setlength{\tabcolsep}{#1}\renewcommand{\arraystretch}{#2}\centering\small}

\newcommand{\gain}[1]{\footnotesize\textcolor{Green}{(+{#1})}}% \textbf

\newcommand{\highlight}[1]{\cellcolor{Gray!16}\textbf{#1}}

\title{\textit{Free-ATM}: Exploring Unsupervised Learning on Diffusion-Generated Images with \\
\textit{Free} \textit{AT}tention \textit{M}asks}
\author{%
   David Junhao Zhang$\textsuperscript{\rm 1}\thanks{Work is partially done during an internship at ByteDance.}$, \quad Mutian Xu$\textsuperscript{\rm 2,4}$, \\
   \textbf{Chuhui Xue}$\textsuperscript{\rm 3}$, \textbf{Wenqing Zhang}$\textsuperscript{\rm 3}$, \textbf{Xiaoguang Han}$\textsuperscript{\rm 2,4}$,  
 \textbf{Song Bai}$\textsuperscript{\rm 3}$, \textbf{Mike Zheng Shou}$\textsuperscript{\rm 1}$\thanks{Corresponding Author.}
 \\
 \\
  $~\textsuperscript{\rm 1}$  Show Lab, National University of Singapore $~\textsuperscript{\rm 2}$  SSE, CUHKSZ  \\
  $~\textsuperscript{\rm 3}$ ByteDance 
  $~\textsuperscript{\rm 4}$ FNii, CUHKSZ
  \\  
}

\begin{document}

\maketitle

\begin{abstract}
Despite the rapid advancement of unsupervised learning in visual representation, it requires training on large-scale datasets that demand costly data collection, and pose additional challenges due to concerns regarding data privacy.
Recently, synthetic images generated by text-to-image diffusion models, have shown great potential for benefiting image recognition.
Although promising, there has been inadequate exploration dedicated to \textit{unsupervised learning} on \textit{diffusion-generated images}.
To address this, we start by uncovering that diffusion models' cross-attention layers inherently provide \textit{annotation-free attention masks} aligned with corresponding text inputs on generated images.
We then investigate the problems of three prevalent unsupervised learning techniques (\textit{i.e.,} contrastive learning, masked modeling, and vision-language pretraining) and introduce customized solutions by fully exploiting the aforementioned free attention masks.
Our approach is validated through extensive experiments that show consistent improvements in baseline models across various downstream tasks, including image classification, detection, segmentation, and image-text retrieval. By utilizing our method, it is possible to close the performance gap between unsupervised pretraining on synthetic data and real-world scenarios.
% Extensive experiments demonstrate that our method consistently improves the baseline models across different downstream tasks such as image classification, detection, and segmentation. Leveraging our approach, the performance gap between unsupervised pretraining on synthetic data and its real-world counterparts can be closed.
% Then we comprehensively investigate the limitations of three prevalent unsupervised learning techniques (\textit{i.e.,} contrastive learning, masked modeling, and vision-language pretraining) and propose tailored frameworks to fully utilize the free attention masks of diffusion-generated images as a free resource to enhance unsupervised learning on synthetic data. 
% We conduct extensive experiments on a host of downstream tasks. The consistent improvement of the baseline models across various downstream tasks (\textit{e.g.,}) demonstrates the effectiveness of our method.

\end{abstract}

\section{Introduction} 
% Training deep neural networks on large amounts of annotated data has achieved great success on a wide range of computer vision tasks \cite{krizhevsky2012imagenet,he2015delving,long2015fully,girshick2015fast,he2016deep,maskrcnn,dosovitskiy2021an,liu2021Swin}. 
% Despite its effectiveness, manually building a sizable labeled dataset is resource-intensive and time-consuming \cite{cifar,imagenet}. 
% In the past few years, \textbf{unsupervised learning} has emerged to overcome this limitation, as it can produce rich representations without any annotations \cite{chen2020contrastive,byol,chen2020simsiam,MoCo,mae,simmim,clip}.
\textbf{Unsupervised learning} is a type of machine learning where models learn to identify patterns or structures in data without explicit labels.
In the past few years, several unsupervised learning techniques have emerged, including contrastive learning \cite{chen2020contrastive,byol,chen2020simsiam,moco}, masked modeling \cite{mae,simmim}, and vision-language pretraining \cite{clip,blip,vilt}, \textit{etc}.
Although these advancements have led to significant progress in visual representation learning, the majority of them rely on pretraining on large-scale datasets, such as ImageNet \cite{imagenet} which contains millions of images.
However, manually building a sizable dataset with decent richness and diversity is often time-consuming and costly. Moreover, present-day concerns about data privacy and usage rights further complicated the acquisition of massive data \cite{privacy}, creating additional obstacles to the development of unsupervised learning.

% Researchers are seeking new ways to address these challenges, such as leveraging \textbf{synthetic images} from generative models.
% First, several efforts have been made in view of using synthetic data for image recognition
To overcome these challenges, using synthetic data for unsupervised pretraining presents itself as a logical option, given its advantageous characteristics such as cost-effectiveness, virtually limitless scalability, enhanced control over data distribution, and improved data privacy and security.
% researchers are seeking alternative ways, such as utilizing \textbf{synthetic images} generated by generative models.

In the computer vision area, there have been some attempts that leverage synthetic data for image recognition tasks. 
Besnier \textit{et al.} \cite{victorthis} and Zhao \textit{et al.} \cite{zhao2022synthesizing} both employ BigGAN \cite{biggan} to produce informative images for training image classifiers. DatasetGAN \cite{zhang21} and BigDatasetGAN \cite{bigDatasetGAN} adopt StyleGAN \cite{stylegan} and BigGAN \cite{biggan} for generating images with pixel-wise labels for segmentation tasks.
% Jahanian \textit{et al.} \cite{jahanian2021generative} alter the latent space of a generator to gain multi-view pairs as the input for contrastive learning.
In addition to using GANs, He \textit{et al.} \cite{he2023is} finds that the revolutionary text-to-image diffusion models such as GLIDE \cite{nichol2022glide} can generate not only high-quality but also diverse images in a customized label space for benefiting image recognition.
This recent study is noteworthy for firstly demonstrating promising results of image understanding using \textbf{diffusion-generated images}.
% which is even better than supervised pretraining.
% Nevertheless, this work solely applies the original MoCo-v2 \cite{MoCov2} to synthetic data without any new designs.
Albeit promising, there has been a lack of in-depth exploration focusing on unsupervised learning on diffusion-generated data.
% advanced frameworks tailored to
We attempt to remedy this defect from the perspective of both diffusion-generated data and unsupervised models.

% Nevertheless, the current investigation of unsupervised learning on synthetic data is still underexplored.
% This motivates us to explore new frameworks tailored to both synthetic data and unsupervised learning models for enjoying the best of both worlds.
% they can complement each other, for 
% to elevate unsupervised learning on synthetic data to the next stage.
% As for synthetic data, although BigDatasetGAN \cite{bigDatasetGAN} shows that pixel-wise labels (\textit{i.e.}, foreground segmentation masks) on synthetic images are helpful for image understanding, it requires manual annotations of a set of images and is constrained in terms of data diversity, as it only produces images of single objects using a class-specific GAN (Fig.~\ref{fig:intro_one}~(a)).

\begin{figure}[t]
    \centering
    \includegraphics[width=1\textwidth]{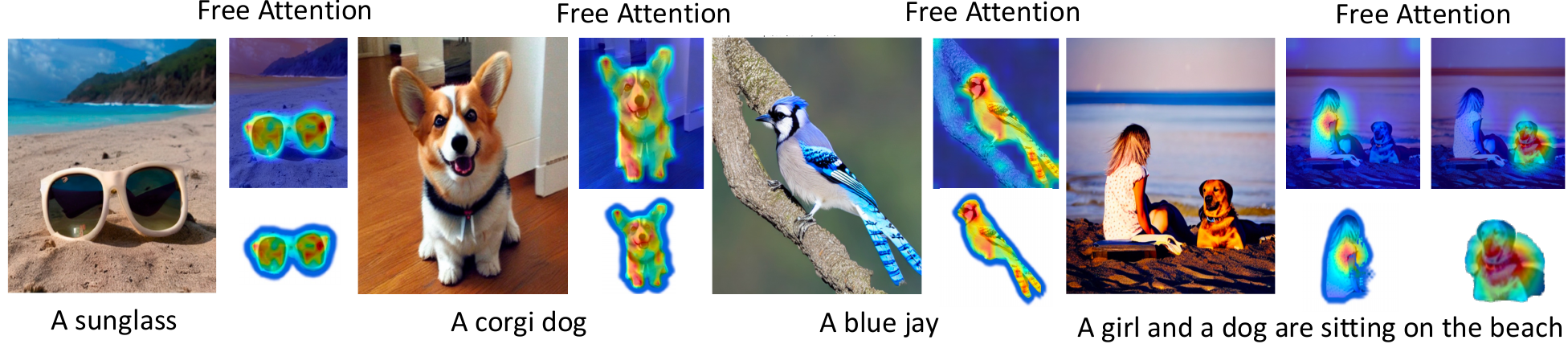}

    \caption{Visualization of diffusion-generated images with freely available attention masks.}
    \label{fig:intro_one}
 
\end{figure}

\textit{i)}~In contrast to class-specific GANs that can only generate images of individual objects, text-to-image diffusion models have the capability to produce diverse images featuring \textit{multiple} objects by utilizing different text tokens.
More importantly, we find that the cross-attention layers of diffusion models naturally provide semantic attention masks aligned with corresponding text inputs on generated images \textit{without} any manual annotations (also indicated in \cite{hertz2022prompt,zhao2023unleashing}), which helps to locate each foreground object as shown in Fig.~\ref{fig:intro_one}.
% making diffusion models a powerful data generator with higher-quality images and annotation-free attention masks.
% Therefore, diffusion models are more powerful synthesis dataset generator given the higher quality images and annotation free attention mask compared with GAN. In this work, we explore the unsupervised pretraining on synthesis images generated by diffusion model.
\textit{ii)}~Next, in view of unsupervised models, we examine the shortcomings of three commonly used unsupervised learning schemes and investigate how the attention masks of diffusion-generated images can be used as a \textit{free} resource to enhance unsupervised learning on synthetic data.
% customized strategies that enjoy the annotation-free attention masks from diffusion generators.

% for model to synthesize datasets of massive high-quality images with semantic mask as shown in Figure. But it still require expensive human annotations to obtain pixel level annotation.)
% Recently, diffusion models exhibit superior text-to-image ability. 
% On the one hand, diffusion model can generate more complex and realstic images than GAN,

\begin{figure}[t]
    \centering
    \includegraphics[width=1\textwidth]{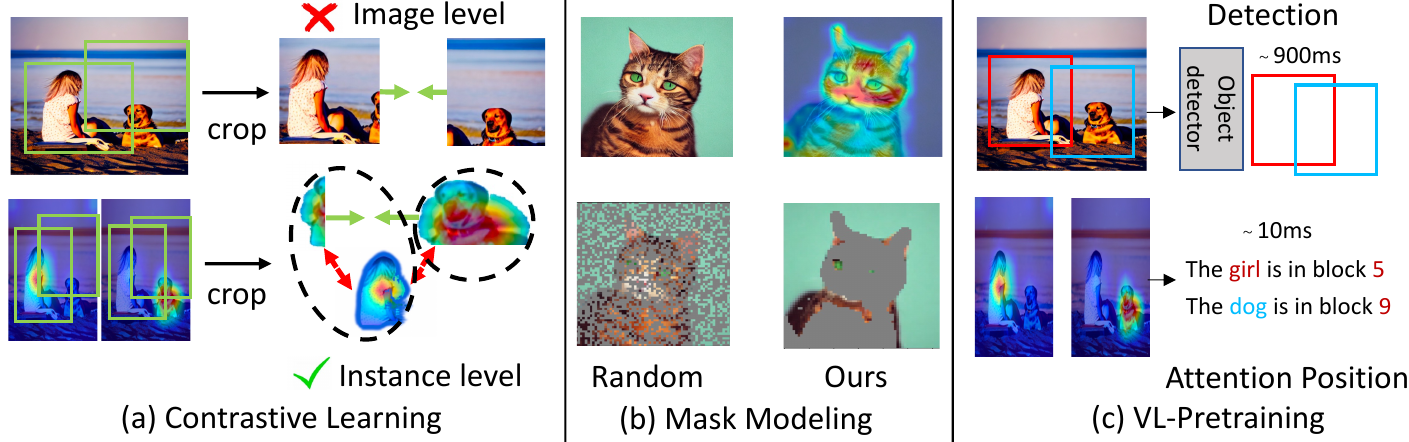}

    \caption{Existing problems in current unsupervised learning frameworks and our solutions.}
    \label{fig:intro_two}

\end{figure}

\textit{Contrastive Learning} (CL) aims to learn representations that can distinguish between positive pairs (similar examples) and negative pairs (dissimilar examples). 
% By minimizing the distance between positive pairs while maximizing the distance between negative pairs. minimize distances between positive pairs and dispel the negative pairs. 
Conventional techniques \cite{chen2020contrastive,byol,chen2020simsiam,moco,mocov2} treat an image with a single object as a complete entity and conduct random crop and augmentations to get positive pairs at the image level.
Yet, such a paradigm is not suitable for diffusion-generated images containing multiple objects.
As discussed earlier, the diversity of diffusion-generated images, which often include multiple instances, presents a significant challenge for traditional random crops in contrastive learning. This is due to the high risk of positive pairs being originated from distinct instances, resulting in ambiguity in model training as the discriminate features of each instance are pulled in (see top row of Fig.~\ref{fig:intro_two} (a)).
To mitigate this issue, we leverage the free attention masks from diffusion generators to ensure that each positive pair comes from the same object. Meanwhile, negative pairs are formed by selecting different instances of images based on their corresponding masks (see bottom row of Fig.~\ref{fig:intro_two} (a)). 
Our approach aids in the acquisition of precise information by the CL model.
% Such instance-level pairs can learn more accurate information than the image-level, considering one image usually contain more than one instance.

\textit{Masked Modeling} (MM) recently achieves widespread success in various vision challenges by reconstructing masked visual patches \cite{mae,simmim,beit}.
A new study \cite{wang2023hard} proposes to identify the patches which are hard to reconstruct, and the study's outcomes reveal that such patches are consistently located within foreground objects. By restoring these particular patches rather than randomly-masked ones, the network can acquire more focused features.
Inspired by this, as indicated in Fig.~\ref{fig:intro_two} (b), we employ the freely available attention map from diffusion generators and present a balanced masking technique for masked modeling that \textit{gradually} increases the masking ratio of foreground object patches. Our strategy enables the MM model to learn both universal and targeted representations.
% Without sophisticated harder patches mining process, we simply mask highest attention score patches to enable network learn more dircriminative  representation.

\textit{Vision-and-Language Pretraining} (VLP) is developed to jointly pretrain visual and language features using image-text matching by learning to align and translate between the two modalities \cite{clip,blip,vilt}. 
% the model enhances its performance across a range of vision-language tasks. 
VLP models predominantly rely on position features, such as those belonging to the objects of interest in an image, to gain a better understanding of the relationships between words and objects.
% To better understand the relationships between words and objects within an image, VLP models dominantly make use of regional features  (\textit{e.g.}, features from objects of interest in an image). 
Nonetheless, locating these features through the use of bounding boxes predicted by object detection algorithms can be a time-consuming process. Additionally, the quality of the position features is heavily dependent on the performance of the object detectors, which may ultimately limit the strength of VLP models.
% Besides, domain-specific object detectors are not feasible for identifying newly-generated data. 
% To address the issue of fine-grained semantic alignment between individual words and objects in the images, some researchers attempt to align the features within bounding boxes with their corresponding words. Yet, vision-language models do not explicitly explore fine-grained semantic alignment between words and objects in the images. Some works attempt to address the problem by aligning features in the bounding box and their corresponding words. However, obtaining a bounding box requires an additional well-trained detection model, which is time-consuming and inefficient.
Fortunately, the attention maps from text-to-image diffusion models naturally align each text prompt with its corresponding object position.
As shown in Fig.~\ref{fig:intro_two} (c), we apply attention masks to supply 
 position information  \textit{without} requiring the extra step of object detection. 
This not only brings greater efficiency to VLP models, but also enhances their overall effectiveness.
% We utilize such benefit and perform simple transform of attention masks into binary masks,
% figure should also have binary masks, how to understand here without refering to fig 3?
% our synthetic images have object positions naturally. This is because we can obtain the attention mask corresponding to each word in the text prompts during text-to-image generation. After simply transforming the attention mask into a binary mask, we know the object lies in which part of the images. The position information is concat with the original text prompt and then to be forced to align visual features during  the pretraining stage.

% Our proposed frameworks elegantly combine
% Extensive experiments are conducted to demonstrate the effectiveness of our method.
Our proposed frameworks, which rely on annotation-\textit{\textbf{Free}} \textit{\textbf{AT}}tention \textit{\textbf{M}}asks from diffusion generators, are collectively referred to as \textit{\textbf{Free-ATM}}.
% which are designed to address three mainstream unsupervised pretraining paradigms on diffusion-generated images. This is achieved by utilizing annotation-free attention masks (\textbf{AT}tention \textbf{M}asks) through our tailored frameworks.
% The proposed frameworks are tailored to three mainstream unsupervised learning paradigms on diffusion-generated images, while leveraging the annotation-free \textbf{AT}tention \textbf{M}asks, so we call our method as \textbf{\textit{Free ATM}}.
% To generate synthetic images with annotation-free attention masks, we adopt a powerful latent diffusion model, Stable Diffusion \cite{latent}. 
% We then proceed to conduct unsupervised pretraining on generated images. %, which are respectively tailored to CL, MM, and VLP.
Extensive experiments show that our method of conducting unsupervised pretraining on diffusion-generated data consistently improves the baseline performance on large-scale real-world benchmarks, including PASVOC \cite{pascal}, COCO \cite{coco}, Cityspace \cite{city}, and ADE20K \cite{ade}, across various downstream tasks like image classification, detection, segmentation, and image-text retrieval.
% We adopt Stable Diffusion \cite{latent} to generate synthetic images with annotation-free attention mask, and conduct unsupervised pretraining under our customized frameworks tailored to CL, MM and
% such as SimCLR \cite{chen2020contrastive}, MoCo-v2 \cite{MoCov2}, MAE \cite{mae} and BLIP \cite{blip} to conduct unsupervised pretraining under 
% conduct extensive experiments to verify the effectiveness of our proposed frameworks.
Besides, our Free-ATM achieves significantly better results compared to simply applying original unsupervised pretraining protocols on diffusion-generated data.
By leveraging Free-ATM, the performance gap between unsupervised pretraining on synthetic data and real-world scenarios can possibly be closed.
Moreover, mixing synthetic data with real-world data for pretraining can further boost performance.
% \mt{more data can bring more power?, which indicates the scalability of our method.}

% Moreover, thanks to our synthesis datasets, the original method can achieve better performance on the popular benchmark without any additional human-annotate and cumbersome data collection process.}
% By utilizing annotate-free attention mask benefit Contrastive Learning, MAE, Vision-Language Pretraining - three popular unsupervised learning framework.
% The code implementations, pretrained models, and synthetic data will all be released upon the acceptance of the paper.

\section{Related Work}

\textbf{Text-to-Image Diffusion Models.} Diffusion models \cite{ddpm,ddim,classifier,diffusiongan,shi2023dragdiffusion} have been making waves in the generative model's landscape, primarily due to their unique approach to generating new data samples. These models commence with a straightforward random noise and progressively denoise it through numerous steps with learned transformations until it mirrors a sample from the desired data distribution. Recently, large-scale text-to-image diffusion models such as Stable Diffusion \cite{latent}, Imagen \cite{imagen}, and GLIDE \cite{nichol2022glide} have made considerable strides and produced striking visual outcomes. Rombach \textit{et al.} proposed an approach where the diffusion process takes place in the latent space, utilizing a UNet \cite{unet} to predict noise and a VAE \cite{vae} decoder to convert the latent feature into pixel space. This approach streamlined the text-to-image diffusion process and accelerated it, becoming a popular choice for diffusion models. Building on the latent diffusion model, Hertz \textit{et al.} \cite{hertz2022prompt} discover that the cross-attention map, which is employed for text-visual interaction in UNet, can accurately represent the foreground object when given the appropriate prompt. They leveraged this characteristic to manipulate images as per requirements by directly modifying the attention map. Similarly, Zhao \textit{et al.} \cite{zhao2023unleashing} use the UNet of the diffusion models as the core structure and extracted the cross-attention map to boost the zero-shot segmentation performance.

In our study, we use the latent diffusion model to generate images in line with the text prompts and to extract the cross-attention map. Using human annotation-free sources, we investigate how attention maps, serving as an additional resource, can enhance unsupervised learning on synthetic images.

\textbf{Synthetic Data from Generative Models.}
Generative Adversarial Networks (GANs) \cite{goodfellow2020generative} have the capability to produce highly realistic and superior quality images. There are numerous studies that utilize GANs to synthesize datasets akin to ImageNet \cite{imagenet}. Li \textit{et al.} \cite{bigDatasetGAN} present BigdatasetGAN, a method that generates a vast amount of images corresponding to ImageNet classes with pixel-level labels, but it necessitates additional human annotation. Jahanian \textit{et al.} \cite{jahanian2021generative} propose a unique method that employs a GAN to generate multiple views of an image, which are then used as positive pairs for contrastive learning. Recently, diffusion models have been gaining precedence in the field of image generation. He \textit{et al.} \cite{he2023is} demonstrate that innovative text-to-image diffusion models, such as GLIDE \cite{nichol2022glide}, can produce data in a custom label space for image recognition. Furthermore, works by Azizi \textit{et al.} \cite{azizi2023synthetic} and Mert Bulent \textit{et al.} \cite{fakeit} employ Imagen \cite{imagen} and Stable Diffusion \cite{latent} to generate images with class labels, thereby improving supervised classification performance. Adding to this, Brandon \textit{et al.} utilize diffusion model inversion to augment images for classifying small datasets.

Rather than focusing on task-specific synthetic data manipulation, we delve into the broader realm of unsupervised learning on synthetic images. 
% This approach can be adapted to a variety of downstream tasks. 
Notably, we leverage the overlooked attention masks, which can offer pixel-level labels without any need for human annotations, thus adding a new dimension to unsupervised learning.

\textbf{Unsupervised Learning.}
\textit{(i) Contrastive Learning} is built upon the foundational idea of drawing positive pairs nearer while distancing negative pairs in the representational space. This method has proven to be effective in learning visual representations without the need for labeled data \cite{bachman2019learning, henaff2019data_CPCv2, wu2018unsupervised, misra2020self_PIRL, oord2018representation, ye2019unsupervised, tian2019contrastive, Wang_2021_CVPR}. A significant advancement in this area is the SimCLR framework \cite{chen2020contrastive}, which has substantially improved the quality of the learned representations via a non-linear transformation head. MoCo \cite{moco}, another impactful work, maintains a memory bank for a vast array of negative samples, and employs a momentum-based method for gentle updates, ensuring improved consistency during learning. \textit{(ii) Masked Modeling} \cite{mae,ibot,data2vec,wmasked,beit}, such as MAE \cite{mae}, SimMIM \cite{simmim}, and iBOT \cite{ibot}, use masked patch reconstruction in combination with basic data augmentation to effectively learn robust representations. \textit{(iii) Vision-Language Pre-training (VLP)}, exemplified by models like CLIP \cite{clip}, BLIP \cite{blip}, and ViLT \cite{vilt}, seeks to boost performance on downstream vision and language tasks by pretraining the model on large-scale image-text pairs to align visual and textual features without any task-specific supervision.
\vspace{-2mm}

In this work, we propose tailored approaches that leverage the freely available attention masks from diffusion generators to enhance the above-mentioned three unsupervised learning frameworks.

\section{Method}
\subsection{Attention Mask from Text-to-Image Diffusion Model}  Latent Text-to-Image Diffusion model \cite{latent} is a generative model that uses a text prompt to create high-quality images through a controlled diffusion process. It first encodes the text prompt into a latent space representation, then uses a diffusion process to gradually transform a noise input into the final image, guided by the encoded text. The model represents the joint understanding of textual and visual data via cross-attention interaction in UNet. Specifically, in the single diffusion step and layer, the text embedding is projected into key as $K\in \mathbb{R}^{L\times C}$ and the visual noise is projected into query as $Q \in \mathbb{R}^{H\times W \times C}$, where $L$ is the text sequence length, $H,W$ are height and width of visual feature and $C$ is the feature dimension. The cross-attention map is achieved by the multiplication of $Q$ and $K$, resulting in
\begin{equation}
A=\text{Softmax}\left(\frac{QK^T}{\sqrt{d}}\right),
\label{equ1}
\end{equation}

where $A\in \mathbb{R}^{H \times W \times L}$ stands as the attention map, illustrating the relationship between textual and visual elements. The attention map $a \in \mathbb{R}^{H \times W}$, corresponding to specific nouns such as `dog' in an $L$ length sentence, is selected. Following this, we compile the attention maps derived from every layer and time step within the diffusion models. These maps are then resized and averaged to form a new map. The visualizations are shown in Fig.~\ref{fig:intro_one}.

\subsection{Prompts Generation} 
We employ the ImageNet \cite{imagenet} label-space as a prompt cue to create synthetic images. It is vital for the model to encounter a wide variety of images in order to learn universal representations applicable to various downstream tasks. A straightforward strategy to diversify synthetic images is to leverage a large language model to transform labels into sentences, effectively augmenting the prompts. However, simply utilizing a large model like T5 \cite{t5}, as suggested by He \textit{et al.}~\cite{he2023is}, could result in unrealistic prompts, thereby producing unrealistic images. This may inadvertently widen the gap between synthetic and real domains, leading to an inferior performance on downstream tasks. To generate more realistic and diverse images, we utilize GPT-3.5-turbo \cite{gpt} to augment prompts. The augmentation templates vary based on the hierarchical level of classes in ImageNet, for example, \textit{“[Class (with  other  class)] is/are  [somewhere]” or “[Class ] with  [other class]  is/are  [doing  something] [somewhere]"}.

\subsection{Utilize Attention Masks for Unsupervised Learning}

Upon generating diverse images and a complimentary attention mask via augmented prompts used as input to diffusion models, we suggest modifications to three common unsupervised learning frameworks: Contrastive Learning, Masked Modeling, and Vision-Language Pretraining. These proposed adjustments fully exploit the attention masks to augment unsupervised learning, thereby enhancing performance on downstream tasks.

% \begin{table}[]
% \centering
% \resizebox{0.65\textwidth}{!}{
% \begin{tabular}{c|c|c}
% \hline
% \hline

% Main Class &Sub-Class & Augment Prompts Templates\\
% \hline
% \multirow{3}{*}{Organism} & Animal      & Class(with other class) is/are doing something somewhere \\
%                           & Plant       & Class(with other class) is/are somewhere                 \\
%                           & People Role & Class is/are doing something somewhere                   \\
%                           \hline
% Artifact                  &    ---         & Adj. class(with other class) is /are somewhere            \\
% \hline
% others                    &    ---         & Adj. class(with other class) is /are somewhere   \\ 
% \hline
% \hline
% \end{tabular}}
% \caption{Prompts augmentation templates}
% \label{prompts}
% \end{table}

\subsubsection{Contrastive Learning} 

\begin{figure}[t]
    \centering
    \includegraphics[width=1\textwidth]{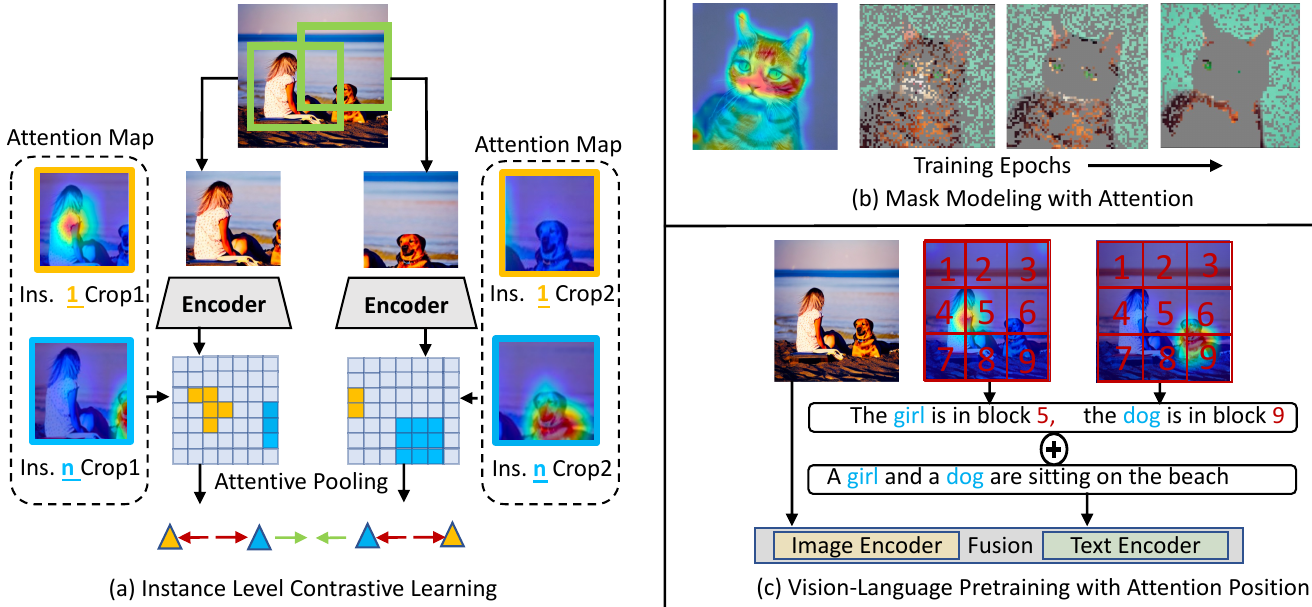}
    \caption{Adaptions for three frameworks to utilize free attention masks.}
    \label{fig:method}
\end{figure}

We adopt two widely used contrastive learning methods - SimCLR \cite{chen2020contrastive} and MoCo-v2 \cite{mocov2}, to incorporate the use of free attention masks. For ease of explanation, we use SimCLR as a representative example, as depicted in Fig. \ref{fig:method}~(a). As previously discussed in the introduction, using image-level features can be problematic when an image contains multiple instances, such as a girl and a dog. The augmented positive pairs, after random cropping, may contain different instances. This could negatively impact network training, as the distinct instance features that should be differentiated are instead conflated. To mitigate this, we use instance features based on the attention mask in place of image features.

Specifically, given an image, we apply a series of augmentations and a random crop, resulting in two cropped images, denoted as $x$ and $x'$. Concurrently, our instance attention masks are cropped in line with the image operation, yielding two sets of masks ${a^{1},\cdots, a^{N}}$, ${a'^{1},\cdots, a'^{N}}$, with each set containing $N$ instances. These two image crops are then input into the encoder (e.g., ResNet50 \cite{he2016deep}), resulting in outputs $z=f(x)$, $z'=f(x')$. At this stage, we utilize the features $z\in \mathbb{R}^{h\times w \times c}, z' \in \mathbb{R}^{h\times w \times c}$, which are derived prior to the last average pooling layer. Next, the $m-th$ instance attention masks $a^{m},a'^{m}$, resized to match the spatial resolution of the encoded features, are mapped to the features $z$ and $z'$, thereby applying attentive pooling. This process results in:

\begin{equation}
z^{m} = \frac{1}{\sum_{i,j} a_{i,j}} \sum_{i,j}^{h,w} a_{i,j}z_{i,j},
\end{equation}

and $z'^{m}\in \mathbb{R}^{C}$.  Following the application of attention pooling, the features $z^{m}$ and $z'^{m}$ are transitioned from the image level to the instance level. Subsequently, a straightforward Multilayer Perceptron (MLP) layer is applied to these features. For instance-level features, we redefine the contrastive loss for $z^{m}$ as

\begin{equation}
l = - \log \frac{\exp(sim( z^{m} \cdot z'^{m}))}{\exp(sim( z^{m} \cdot z'^{m}))+{\sum_{n=1}^{N}}_{[m \neq n]} \exp( sim( z^m \cdot z^{n}))},
\end{equation}
\label{con}

where $sim\left(z^m, z'^m\right)=\frac{{z^{m}}^{\top} \cdot z'^m}{\left|z^m\right| \cdot\left|z'^m\right|}$. In this loss computation, we consider the features of the same instance from the two crops as the positive pair, and the features of different instances as negative pairs. In Equation (3), $N$ signifies the total number of instances in the image. When extended to the batch dimension, $N$ can represent the total number of instances in the batch.

For MoCo-v2, the encoders for the two image crops are distinct. One encoder is updated by the Exponential Moving Average (EMA) \cite{moco} of the other encoder. Furthermore, instance-level features, as opposed to image-level, are updated and stored in the memory bank. In addition to addressing the issue where positive pairs may comprise different instances, this strategy enables every image to provide a wealth of information. This greatly aids the network and allows for the learning of a more diverse representation, given the fact that each image typically encompasses multiple instances.

\subsubsection{Masked Modeling} 

As indicated in \cite{wang2023hard}, certain image patches can be challenging to reconstruct, and these often represent the foreground objects in images. Prioritizing the reconstruction of these difficult patches aids the network in learning a more discriminative representation. To accomplish this, Wang \textit{et al.} \cite{wang2023hard} introduce an additional teacher-student network to predict these difficult patches. However, deploying an extra teacher-student model will bring extra computation costs and complicate the learning process. In contrast, our free attention mask naturally embodies the importance score of the foreground object mask. The scores of the attention map, ranging from low to high, correspond to patches from easy to difficult. This allows us to discard the teacher-student model for identifying challenging patches.

An intuitive approach would be to mask the patches with the highest attention map scores and then use pretraining from scratch to reconstruct the masked images. However, solely focusing on reconstructing the difficult patches may cause the network to overly concentrate on the foreground object, which could be detrimental to learning a more universal representation of the entire image. To mitigate this, during the initial stages of training, we continue to mask the patches randomly. As the training epochs increase (Fig. \ref{fig:method} (b)), we gradually raise the ratio of masked patches determined by the highest importance scores, and reduce the ratio of randomly selected masked patches. This balanced approach enables the network to learn both universal and targeted representations simultaneously.

\subsubsection{Vision-Language Pretraining.}

The capacity for position grounding \cite{glip,loctex} is vital for a vision-language model to excel in cross-modality downstream tasks. Some studies strive to enhance this capability by incorporating bounding box and region features as additional visual inputs during vision-language pretraining. However, obtaining these features and bounding boxes for the objects in the image necessitates the use of a robust, pre-trained offline detection model, such as Fast-RCNN \cite{girshick2015fast}. This process can be time-consuming and often results in a significant increase in the parameters for the vision-language models. In contrast, our synthetic data inherently includes the bounding box of the object (nouns). This is made possible as we can readily transform the attention mask into a binary mask, with pixels marked as `1' representing the foreground region, thereby allowing us to obtain the bounding box. 

Rather than extracting regions using bounding boxes as inputs for the visual encoder, we employ position-aware prompts, inspired by \cite{position}, which does not impose additional parameters or computational demands on the vision-language model. As illustrated in Fig.~\ref{fig:method} (c), we initially divide the image into $N$ blocks. After determining the bounding box of the object, we can identify the block in which the object's center is located. Using this information, we generate position-aware prompts following the template:  “\textit{\textit{ The [\textcolor{ocolor}{O}]} is in block [\textcolor{pcolor}{P}].}”

% \begin{equation}
% ``\textit{\textit{ The [\textcolor{ocolor}{O}]} is in   block [\textcolor{pcolor}{P}].}"
% \end{equation}

Subsequently, we concatenate these prompts for all objects in the images with the original prompt. This forms the input for the text encoder, enhancing the position-grounding ability of the vision-language model. The model in focus, BILP, is adapted to our needs for Vision-Language Pretraining (VLP). We then conduct end-to-end training using conventional objectives. Consistent with the methodologies outlined in \cite{blip,vilt,itm}, the training process involves the use of Language Modeling (LM) loss, Image-Text Matching (ITM) loss, and Image-Text Contrastive (ITC) loss. Note that the object's positional information is only required during the pre-training stage. For downstream tasks, we evaluate the model using standard end-to-end methods, without the need for object information.

\section{Experiments}
In the process of pretraining for contrastive learning and masked modeling, we generate approximately 1.2 million images using augmented prompts, a quantity that matches precisely the original ImageNet-1K \cite{imagenet} dataset for fair comparisons. For the supervised pretraining phase, we utilize real data tagged with class labels. As for pretraining in the vision-language task, we select a subset from CC3M \cite{sharma2018conceptual}, which encompasses 0.3 million image-text pairs, and employ the original captions to generate images.

\subsection{Contrastive Learning}

\subsubsection{SimCLR}

\textbf{Pretraining.} We employ ResNet50~\cite{He_2016_CVPR} as the encoder for our pretraining. To maintain fairness in our comparisons, we adhere to the same 200 pretraining epochs across all settings, and all hyperparameters for training are aligned with those outlined in the original SimCLR paper\cite{chen2020contrastive}. As outlined in Tab. \ref{tab:simclr}, various terms denote different pretraining methods. `Random init' implies no pretraining, 'supervised' refers to pretraining with ImageNet$\&$labels, and `real' indicates self-supervised pretraining on ImageNet. `Synthetic' stands for self-supervised pretraining on purely synthetic images, while `synthetic w/ ours' signifies our adapted method using synthetic images with free attention masks. `Mix' involves a blend of all masked synthetic and real images.
 
\textbf{Object Detection and  Segmentation.}
The pretrained network is utilized to initialize our feature extractor. For object detection and instance segmentation in COCO \cite{coco}, we, in accordance with \cite{moco}, modify the Mask-RCNN \cite{He_2017_ICCV} to be equipped with feature pyramid networks. The complete model is fine-tuned on the training dataset, following a standard 1x schedule (12 epochs), and we report the results as bounding box AP ($AP^{b}$) and instance mask AP($AP^{m}$). In the case of semantic segmentation for Cityspace \cite{city}, we fine-tune the model over a span of 160 epochs, reporting the results in terms of $mIoU$ (mean Intersection over Union).

\textbf{Results.}
As shown in Tab. \ref{tab:simclr}, self-supervised pretraining using purely synthetic data leads to a noticeable performance discrepancy on the COCO and City Space datasets when compared to the use of purely real data. However, with the introduction of the free attention mask, we observe a marked improvement in results: an increase of 1.3$\%$ and 1.1$\%$ on the COCO dataset, and 0.8$\%$ on the CitySpace dataset. This improvement narrows the gap with real data to an almost negligible level. These findings underscore the efficacy of our approach to instance-level contrastive learning, facilitated by the use of free attention masks. 
Moreover, when combining both real and synthetic data, we see not only an enhancement in performance but also a surpassing of the results achieved using purely real data. This is accomplished without incurring any costs associated with human annotation or data collection.

\subsubsection{MoCo-v2}
\textbf{Pretraining.} We employ ResNet-50 as the encoder and train it for a total of 200 epochs. During this training phase, we adopt a learning rate adjustment strategy wherein the learning rate is multiplied by 0.1 at two specific points: the 120th and the 160th epochs. This adjustment is applied consistently across all experimental settings. Aside from these specifics, all other training hyperparameters strictly adhere to the guidelines and recommendations provided in the original MoCo-v2 paper~\cite{mocov2}. This approach ensures that our training process is rigorous and consistent with established best practices.

\begin{table}
   \begin{minipage}{0.48\textwidth}
     
    \caption{ SimCLR~\cite{chen2020contrastive} downstream results.}
   \centering
      \tablestyle{2.6pt}{1.38}
      \begin{tabular}{l|cc|c}
   \toprule[1pt]
           & \multicolumn{2}{c|}{COCO} & Cityspace \\
                  &    $AP^b$          & $AP^m$     & $mIoU$     \\
                  \hline
                
random ini        & 31.4        & 28.5       & 65.2      \\
supervised        & \highlight{39.2}        & \highlight{35.5}       & 74.2      \\
real              & 37.7        & 34.2       & 74.8      \\
\hline
synthetic         & 36.4        & 33.1       & 73.7      \\
synthetic w/ ours & 37.7 \gain{1.3}        & 34.2 \gain{1.1}      & 74.5\gain{0.8}      \\
mix w/ ours       & 38.7          & 34.8       & \highlight{75.0}       \\
   \toprule[1pt] 
\end{tabular}

         \label{tab:simclr}
      \end{minipage}
      \hspace{5mm}
        \begin{minipage}{0.48\textwidth}
    
      \caption{ BLIP~\cite{blip} image-text retrieval results.}
      \centering
      \tablestyle{2.6pt}{1.38}
      \begin{tabular}{l|cc}
      \toprule[1pt]

          & \multicolumn{2}{c}{MS-COCO} \\
   
          & $tr@1$        & $ir@1$         \\
       \hline
real      & 58.1         & 44.2 \\
\hline

VQGAN \cite{esser2021taming} & 35.6         & 32.0         \\
DALL-E2 \cite{ramesh2021zero}   & 44.5        & 38.6         \\
\hline
Stable \cite{latent}    & 52.3         & 40.9         \\
w/ ours & 54.9  \gain{2.6}       & 43.8 \gain{2.9}         \\
mix w/ ours      &\highlight{60.8}          & \highlight{46.2} \\
   \toprule[1pt]
\end{tabular}

         \label{tab:vl}
        %  \vspace{-0.1cm}  
      \end{minipage}
   \hfill
 
   \begin{minipage}{0.48\textwidth}
     \vspace{3mm}
     \caption{MoCo-v2~\cite{mocov2} downstream results.}

   \centering
      \tablestyle{2.6pt}{1.2}
       \begin{tabular}{l|c|c|c}
\toprule[1pt]
                  & PASVOC & COCO & Citysapce \\
                  
                  & $AP^{b}_{50}$  & $AP^{b}$ & $mIoU$     \\
                  \hline
random ini        & 59.0   & 31.4 & 65.2      \\
supervised        & 81.6   & 39.2 & 74.2      \\
real              & 82.4   & 39.8 & 75.0
\\
\hline
synthetic         & 81.6   & 37.9 & 74.1      \\
synthetic w/ ours & 82.1 \gain{0.5}    & 38.3 \gain{0.4} & 74.8 \gain{0.7}      \\
mix w/ ours             & \highlight{82.6}   & \highlight{40.1} & \highlight{75.3} \\
\toprule[1pt]
\end{tabular}

         \label{tab: moco}
      \end{minipage}
      \hspace{5mm}
      \vspace{6mm}
      \begin{minipage}{0.48\textwidth}
          \vspace{3mm}
      \caption{MAE~\cite{mae} downstream results.}
      \centering
      \tablestyle{2.6pt}{1.38}
     \begin{tabular}{l|c|c}
\toprule[1pt]
                  & ImageNet  & ADE20K \\
                  & $acc$ &$mIoU$  \\
                  \hline
supervised        & 81.0      & 47.4   \\
real              & 83.6      & 48.1   \\
\hline
synthetic         & 82.7     & 47.6   \\
synthetic w/ ours & 83.2 \gain{0.5}     & 48.0 \gain{0.4}  \\
mix w/ ours             & \highlight{83.8}      & \highlight{48.5}  \\
\toprule[1pt]
\end{tabular}

         \label{tab:mae}
      \end{minipage}

  \end{table}

\textbf{Object Detection and Semantic Segmentation.}
In the context of object detection evaluation, we adhere to the standard protocol of fine-tuning a Faster R-CNN detector (with a C4 backbone) on the VOC trainval07+12 set, as per the 2x schedule mentioned in \cite{Wang_2021_CVPR}. The evaluation is then carried out on the VOC test2007 set. In evaluating COCO detection, our fine-tuning strategies remain consistent with those applied in the previously discussed SimCLR framework. When it comes to semantic segmentation in Cityspace, we utilize an FCN \cite{long2015fully} model. This model is trained on the train-fine dataset over 40k iterations. The testing is then carried out on the validation dataset.

\textbf{Results.}
Tab. \ref{tab: moco} reveals a pattern similar to that observed with SimCLR: when pretraining is conducted solely on synthetic images, there is a noticeable performance gap compared to pretraining on real images. Yet, interestingly, the results obtained using only synthetic images can rival and even surpass those achieved through supervised learning on real images, a process that typically requires substantial human annotation and collection efforts. Furthermore, the application of free attention masks enhances the performance on synthetic images to a level comparable with that on real data, effectively bridging the gap between synthetic and real images. When pretraining incorporates both synthetic and real images, the results show an improvement over both supervised learning with real data and self-supervised pretraining, all without the need for human labor.

\subsection{Masked Modeling}

\textbf{Pretraining.}
We employ MAE \cite{mae}, a representative masked modeling method.
We follow its original pretraining protocol for the ViT-B \cite{vit}, spanning a total of 1600 epochs. The overall mask ratio is set at 75$\%$. Additionally, we adopt a strategy to incrementally increase the ratio $\beta$ of masked patches according to the highest attention score, with a ceiling of 0.8, following a linear progression as epochs increasing. Thus, a portion of the masked patches, specifically ($\beta \times 0.75$) of them, are determined based on the attention score. The remaining masked patches, which account for ($(1-\beta) \times 0.75$) of the total, are selected at random.

\textbf{Image Classification and Semantic Segmentation.}
For classification tasks, we utilize the Masked Auto Encoder (MAE) \cite{mae}. We carry out end-to-end fine-tuning for 100 epochs on the ImageNet-1K training set and subsequently report the top-1 accuracy on the validation set. For segmentation tasks, we use UperNet \cite{uper} as the segmentation head. This is then subjected to end-to-end fine-tuning on the ADE20K \cite{ade} dataset for a total of 160k iterations.

\textbf{Results.}
As shown in Tab. \ref{tab:mae}, MAE \cite{mae} achieves consistent trend results when compared to MoCo-v2. Utilizing  free attention masks can assist in bridging the discrepancy between synthetic and real data. Additionally, by exclusively employing synthetic data, which does not require human annotation or collection costs, we can significantly outperform fully annotated supervised training (83.8 \textit{vs.} 81.0).

\subsection{Vision Language Pretraining}
\textbf{Pretraining.}
We train the vision-language model following BLIP~\cite{blip} on 0.3 M text-image pairs until the loss converges.

\textbf{COCO-Retrieval.}
We evaluate BLIP on the COCO datasets, focusing on image-to-text retrieval (TR) and text-to-image retrieval (IR). The pre-trained model is fine-tuned using both ITC and ITM loss functions. Additionally, we implement a re-ranking strategy to further refine the retrieval results.

\textbf{Results.}
Tab. \ref{tab:vl} illustrates that with free attention mask, which provides position-aware prompts, enables the model to attain a substantial improvement over results obtained using purely synthetic data. Furthermore, combining synthetic data with the mask and real data considerably enhances the results, all without incurring any costs associated with human annotation or data collection.
\subsection{Ablation Study}
\textbf{Image quality.}
Stable Diffusion is the most effective open-source model for text-to-image generation, surpassing both VQGAN \cite{esser2021taming} and DALL-E2-LAION \cite{ramesh2021zero} in terms of visual quality. We have compared the impact of pretraining on images synthesized by these three generators. Tab. \ref{tab:vl} reveals that higher image quality correlates with improved results. This suggests that as more powerful generators emerge in the future, the prospect of pretraining on synthetic data becomes increasingly promising.
\begin{wraptable}{r}{4.3cm}
	\centering
 
 \label{tab: p-rompts}
	\begin{tabular}{ccc}
 \toprule[1pt]
     PASVOC & base & augment\\
       \hline
       $AP^{50}$& 81.0  & \highlight{81.6}\\
         \toprule[1pt]

	\end{tabular}
  
 \caption{Prompts ablations.}
 \vspace{-4mm}
\end{wraptable}

\textbf{Prompts design.}
We assess the significance of augmented prompts in our study, simply employing a basic prompt, "a photo of [class]," for comparison purposes. As indicated in Tab. 5, augmented prompts contribute significantly to the generation of a broader range of images, enhancing the pretraining process.

\textbf{More data, more power.}
We examine the effectiveness of our framework powered by an increasing quantity of diffusion-generated images.
Here we adopt MoCo-v2 \cite{mocov2} to incorporate our Free-ATM, and report the pretraining transfer performance on the PASVOC \cite{pascal} dataset.
The results, displayed in Tab.~\ref{tab: numbers}, demonstrate a consistent performance enhancement in line with the increase in synthetic image count. This observation suggests that synthetic images can indeed scale in a manner similar to real images.
\begin{table}[h]
\centering
\begin{tabular}{ccccc}
\toprule[1pt]

 number of synthetic images & 0.5M  & 1M &1.5M &2M \\

  \hline 
  $AP^{b}_{50}$&81.3 &82.1 &82.4 &82.7

\\

\toprule[1pt]
\end{tabular}
\label{tab: numbers}

\caption{Using more synthetic images for our unsupervised pretraining framework.}
\end{table}

Moreover, by utilizing 1.5M synthetic images, our approach achieves equivalent performance to the original MoCo-v2, which is trained on 1M real-world images from ImageNet \cite{imagenet}. 
This finding highlights that by increasing the amount of synthetic data, our method can effectively close the performance gap between unsupervised pretraining on synthetic data and real-world scenarios.
% This shows that by using more synthetic images, we can close the performance gap between unsupervised pretraining on synthetic data and real-world scenarios.
Supported by our Free-ATM, synthetic data can be used as a viable alternative to real-world data for unsupervised pretraining, which can be beneficial in scenarios where real-world data is scarce or insufficient.
In the future, we will keep exploring the upper bounds of our framework using more synthetic data.

\section{More Visualizations of Attention Masks}

\begin{figure}[h]
    \centering
    \includegraphics[width=1\textwidth]{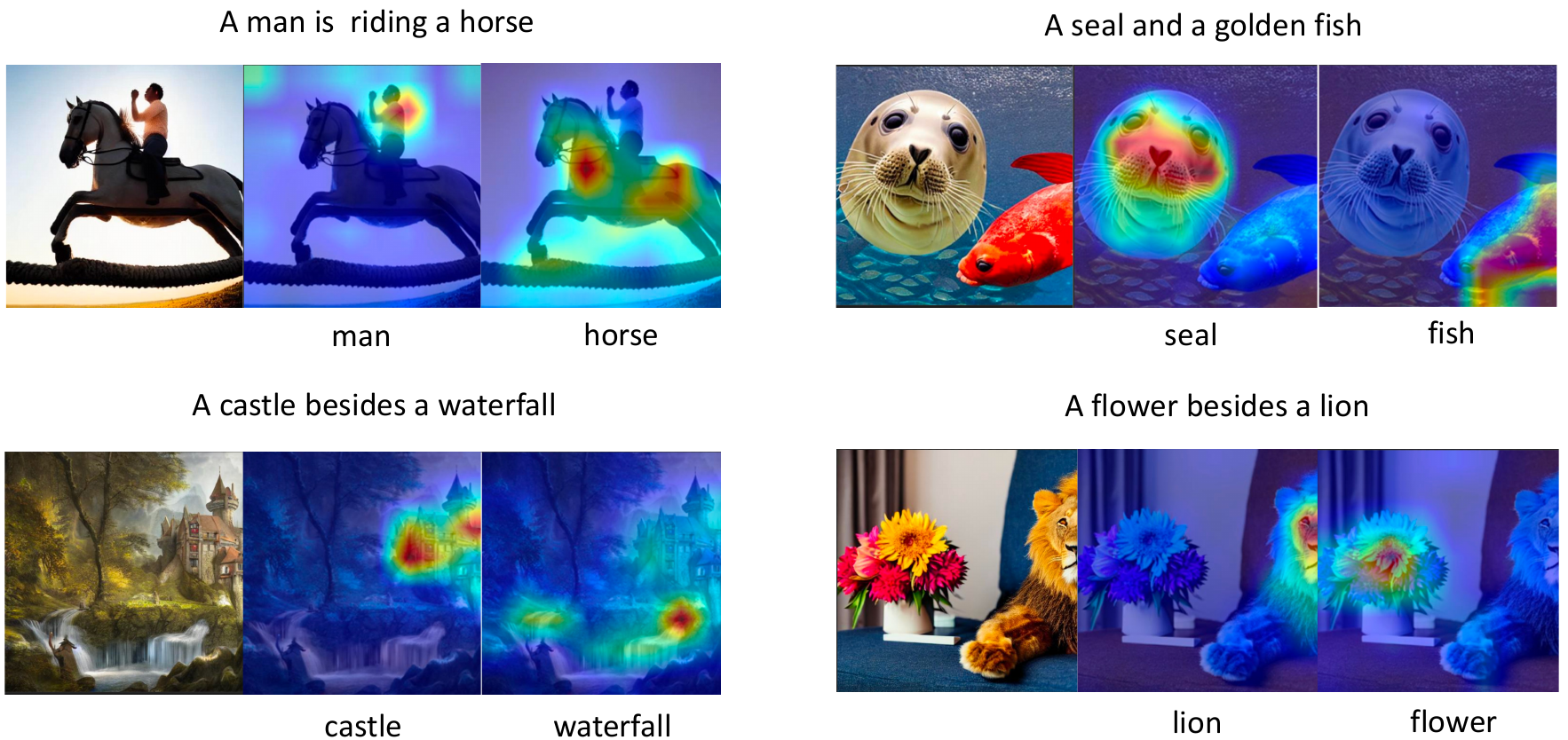}
    \caption{Visualizations of Attention Masks.}
    \label{fig:supp}
\end{figure}

Fig. \ref{fig:supp} provides diverse visualizations of the freely-available attention masks from text-to-image diffusion models, where the cross-attention maps between object instances and their corresponding text prompts are decently aligned. 
% \begin{abstract}
%   The abstract paragraph should be indented \nicefrac{1}{2}~inch (3~picas) on
%   both the left- and right-hand margins. Use 10~point type, with a vertical
%   spacing (leading) of 11~points.  The word \textbf{Abstract} must be centered,
%   bold, and in point size 12. Two line spaces precede the abstract. The abstract
%   must be limited to one paragraph.
% \end{abstract}

\section{Conclusion}
We have proposed Free-ATM, a technique that utilizes the readily available attention masks from diffusion generators,  to improve unsupervised learning on diffusion-generated images. Our method is demonstrated to be effective through extensive experimental results across various downstream tasks and benchmarks.
We anticipate that our approach will provide new directions for future research on leveraging synthetic images to tackle computer vision problems.

% \section*{References}
{\small
\bibliography{egbib}

\begin{thebibliography}{10}
\providecommand{\url}[1]{#1}
\csname url@samestyle\endcsname
\providecommand{\newblock}{\relax}
\providecommand{\bibinfo}[2]{#2}
\providecommand{\BIBentrySTDinterwordspacing}{\spaceskip=0pt\relax}
\providecommand{\BIBentryALTinterwordstretchfactor}{4}
\providecommand{\BIBentryALTinterwordspacing}{\spaceskip=\fontdimen2\font plus
\BIBentryALTinterwordstretchfactor\fontdimen3\font minus
  \fontdimen4\font\relax}
\providecommand{\BIBforeignlanguage}[2]{{%
\expandafter\ifx\csname l@#1\endcsname\relax
\typeout{** WARNING: IEEEtran.bst: No hyphenation pattern has been}%
\typeout{** loaded for the language `#1'. Using the pattern for}%
\typeout{** the default language instead.}%
\else
\language=\csname l@#1\endcsname
\fi
#2}}
\providecommand{\BIBdecl}{\relax}
\BIBdecl

\bibitem{chen2020contrastive}
T.~Chen, S.~Kornblith, M.~Norouzi, and G.~Hinton, ``A simple framework for
  contrastive learning of visual representations,'' in \emph{ICML}, 2020.

\bibitem{byol}
J.-B. Grill, F.~Strub, F.~Altch\'{e}, C.~Tallec, P.~Richemond, E.~Buchatskaya,
  C.~Doersch, B.~Avila~Pires, Z.~Guo, M.~Gheshlaghi~Azar, B.~Piot,
  k.~kavukcuoglu, R.~Munos, and M.~Valko, ``Bootstrap your own latent - a new
  approach to self-supervised learning,'' in \emph{NeurIPS}, 2020.

\bibitem{chen2020simsiam}
X.~Chen and K.~He, ``Exploring simple siamese representation learning,'' in
  \emph{CVPR}, 2021.

\bibitem{moco}
K.~He, H.~Fan, Y.~Wu, S.~Xie, and R.~Girshick, ``Momentum contrast for
  unsupervised visual representation learning,'' in \emph{CVPR}, 2020.

\bibitem{mae}
K.~He, X.~Chen, S.~Xie, Y.~Li, P.~Doll{\'a}r, and R.~Girshick, ``Masked
  autoencoders are scalable vision learners,'' in \emph{CVPR}, 2022.

\bibitem{simmim}
Z.~Xie, Z.~Zhang, Y.~Cao, Y.~Lin, J.~Bao, Z.~Yao, Q.~Dai, and H.~Hu, ``Simmim:
  A simple framework for masked image modeling,'' in \emph{CVPR}, 2022.

\bibitem{clip}
A.~Radford, J.~W. Kim, C.~Hallacy, A.~Ramesh, G.~Goh, S.~Agarwal, G.~Sastry,
  A.~Askell, P.~Mishkin, J.~Clark \emph{et~al.}, ``Learning transferable visual
  models from natural language supervision,'' in \emph{ICML}, 2021.

\bibitem{blip}
J.~Li, D.~Li, C.~Xiong, and S.~Hoi, ``Blip: Bootstrapping language-image
  pre-training for unified vision-language understanding and generation,'' in
  \emph{ICML}, 2022.

\bibitem{vilt}
W.~Kim, B.~Son, and I.~Kim, ``Vilt: Vision-and-language transformer without
  convolution or region supervision,'' in \emph{ICML}, 2021.

\bibitem{imagenet}
J.~Deng, W.~Dong, R.~Socher, L.-J. Li, K.~Li, and L.~Fei-Fei, ``Imagenet: A
  large-scale hierarchical image database,'' in \emph{CVPR}, 2009.

\bibitem{privacy}
T.~Orekondy, B.~Schiele, and M.~Fritz, ``Towards a visual privacy advisor:
  Understanding and predicting privacy risks in images,'' in \emph{ICCV}, 2017.

\bibitem{victorthis}
V.~Besnier, H.~Jain, A.~Bursuc, M.~Cord, and P.~Pérez, ``This dataset does not
  exist: Training models from generated images,'' in \emph{ICASSP}, 2020.

\bibitem{zhao2022synthesizing}
B.~Zhao and H.~Bilen, ``Synthesizing informative training samples with gan,''
  \emph{NeurIPS 2022 Workshop on Synthetic Data for Empowering ML Research},
  2022.

\bibitem{biggan}
A.~Brock, J.~Donahue, and K.~Simonyan, ``Large scale {GAN} training for high
  fidelity natural image synthesis,'' in \emph{ICLR}, 2019.

\bibitem{zhang21}
Y.~Zhang, H.~Ling, J.~Gao, K.~Yin, J.-F. Lafleche, A.~Barriuso, A.~Torralba,
  and S.~Fidler, ``Datasetgan: Efficient labeled data factory with minimal
  human effort,'' in \emph{CVPR}, 2021.

\bibitem{bigDatasetGAN}
D.~Li, H.~Ling, S.~W. Kim, K.~Kreis, A.~Barriuso, S.~Fidler, and A.~Torralba,
  ``Bigdatasetgan: Synthesizing imagenet with pixel-wise annotations,'' in
  \emph{CVPR}, 2022.

\bibitem{stylegan}
T.~Karras, S.~Laine, and T.~Aila, ``A style-based generator architecture for
  generative adversarial networks,'' \emph{TPAMI}, 2021.

\bibitem{he2023is}
R.~He, S.~Sun, X.~Yu, C.~Xue, W.~Zhang, P.~Torr, S.~Bai, and X.~QI, ``{IS}
  {SYNTHETIC} {DATA} {FROM} {GENERATIVE} {MODELS} {READY} {FOR} {IMAGE}
  {RECOGNITION}?'' in \emph{ICLR}, 2023.

\bibitem{nichol2022glide}
A.~Nichol, P.~Dhariwal, A.~Ramesh, P.~Shyam, P.~Mishkin, B.~McGrew,
  I.~Sutskever, and M.~Chen, ``Glide: Towards photorealistic image generation
  and editing with text-guided diffusion models,'' in \emph{ICML}, 2022.

\bibitem{hertz2022prompt}
A.~Hertz, R.~Mokady, J.~Tenenbaum, K.~Aberman, Y.~Pritch, and D.~Cohen-Or,
  ``Prompt-to-prompt image editing with cross attention control,'' \emph{arXiv
  preprint arXiv:2208.01626}, 2022.

\bibitem{zhao2023unleashing}
W.~Zhao, Y.~Rao, Z.~Liu, B.~Liu, J.~Zhou, and J.~Lu, ``Unleashing text-to-image
  diffusion models for visual perception,'' \emph{arXiv preprint
  arXiv:2303.02153}, 2023.

\bibitem{mocov2}
X.~Chen, H.~Fan, R.~Girshick, and K.~He, ``Improved baselines with momentum
  contrastive learning,'' \emph{arXiv preprint arXiv:2003.04297}, 2020.

\bibitem{beit}
H.~Bao, L.~Dong, S.~Piao, and F.~Wei, ``{BE}it: {BERT} pre-training of image
  transformers,'' in \emph{ICLR}, 2022.

\bibitem{wang2023hard}
H.~Wang, K.~Song, J.~Fan, Y.~Wang, J.~Xie, and Z.~Zhang, ``Hard patches mining
  for masked image modeling,'' in \emph{CVPR}, 2023.

\bibitem{pascal}
M.~Everingham, L.~Van~Gool, C.~K. Williams, J.~Winn, and A.~Zisserman, ``The
  pascal visual object classes (voc) challenge,'' \emph{IJCV}, 2010.

\bibitem{coco}
T.-Y. Lin, M.~Maire, S.~Belongie, J.~Hays, P.~Perona, D.~Ramanan,
  P.~Doll{\'a}r, and C.~L. Zitnick, ``Microsoft coco: Common objects in
  context,'' in \emph{ECCV}, 2014.

\bibitem{city}
M.~Cordts, M.~Omran, S.~Ramos, T.~Rehfeld, M.~Enzweiler, R.~Benenson,
  U.~Franke, S.~Roth, and B.~Schiele, ``The cityscapes dataset for semantic
  urban scene understanding,'' in \emph{CVPR}, 2016.

\bibitem{ade}
B.~Zhou, H.~Zhao, X.~Puig, T.~Xiao, S.~Fidler, A.~Barriuso, and A.~Torralba,
  ``Semantic understanding of scenes through the ade20k dataset,'' \emph{IJCV},
  2019.

\bibitem{ddpm}
J.~Ho, A.~Jain, and P.~Abbeel, ``Denoising diffusion probabilistic models,'' in
  \emph{NeurIPS}, 2020.

\bibitem{ddim}
J.~Song, C.~Meng, and S.~Ermon, ``Denoising diffusion implicit models,'' in
  \emph{ICLR}, 2021.

\bibitem{classifier}
J.~Ho and T.~Salimans, ``Classifier-free diffusion guidance,'' \emph{arXiv
  preprint arXiv:2207.12598}, 2022.

\bibitem{diffusiongan}
P.~Dhariwal and A.~Nichol, ``Diffusion models beat gans on image synthesis,''
  in \emph{NeurIPS}, 2021.

\bibitem{shi2023dragdiffusion}
Y.~Shi, C.~Xue, J.~Pan, W.~Zhang, V.~Y. Tan, and S.~Bai, ``Dragdiffusion:
  Harnessing diffusion models for interactive point-based image editing,''
  \emph{arXiv preprint arXiv:2306.14435}, 2023.

\bibitem{latent}
R.~Rombach, A.~Blattmann, D.~Lorenz, P.~Esser, and B.~Ommer, ``High-resolution
  image synthesis with latent diffusion models,'' in \emph{CVPR}, 2022.

\bibitem{imagen}
C.~Saharia, W.~Chan, S.~Saxena, L.~Li, J.~Whang, E.~L. Denton, K.~Ghasemipour,
  R.~Gontijo~Lopes, B.~Karagol~Ayan, T.~Salimans \emph{et~al.},
  ``Photorealistic text-to-image diffusion models with deep language
  understanding,'' in \emph{NeurIPS}, 2022.

\bibitem{unet}
O.~Ronneberger, P.~Fischer, and T.~Brox, ``U-net: Convolutional networks for
  biomedical image segmentation,'' in \emph{MICCAI}, 2015.

\bibitem{vae}
D.~P. Kingma and M.~Welling, ``Auto-encoding variational bayes,'' \emph{arXiv
  preprint arXiv:1312.6114}, 2013.

\bibitem{goodfellow2020generative}
I.~Goodfellow, J.~Pouget-Abadie, M.~Mirza, B.~Xu, D.~Warde-Farley, S.~Ozair,
  A.~Courville, and Y.~Bengio, ``Generative adversarial networks,''
  \emph{Communications of the ACM}, 2020.

\bibitem{jahanian2021generative}
A.~Jahanian, X.~Puig, Y.~Tian, and P.~Isola, ``Generative models as a data
  source for multiview representation learning,'' in \emph{ICLR}, 2022.

\bibitem{azizi2023synthetic}
S.~Azizi, S.~Kornblith, C.~Saharia, M.~Norouzi, and D.~J. Fleet, ``Synthetic
  data from diffusion models improves imagenet classification,'' \emph{arXiv
  preprint arXiv:2304.08466}, 2023.

\bibitem{fakeit}
M.~B. Sariyildiz, K.~Alahari, D.~Larlus, and Y.~Kalantidis, ``Fake it till you
  make it: Learning transferable representations from synthetic imagenet
  clones,'' in \emph{CVPR}, 2023.

\bibitem{bachman2019learning}
P.~Bachman, R.~D. Hjelm, and W.~Buchwalter, ``Learning representations by
  maximizing mutual information across views,'' in \emph{NeurIPS}, 2019.

\bibitem{henaff2019data_CPCv2}
O.~J. H{\'e}naff, A.~Srinivas, J.~De~Fauw, A.~Razavi, C.~Doersch, S.~Eslami,
  and A.~v.~d. Oord, ``Data-efficient image recognition with contrastive
  predictive coding,'' \emph{arXiv preprint arXiv:1905.09272}, 2019.

\bibitem{wu2018unsupervised}
Z.~Wu, Y.~Xiong, S.~X. Yu, and D.~Lin, ``Unsupervised feature learning via
  non-parametric instance discrimination,'' in \emph{CVPR}, 2018.

\bibitem{misra2020self_PIRL}
I.~Misra and L.~v.~d. Maaten, ``Self-supervised learning of pretext-invariant
  representations,'' in \emph{CVPR}, 2020.

\bibitem{oord2018representation}
A.~v.~d. Oord, Y.~Li, and O.~Vinyals, ``Representation learning with
  contrastive predictive coding,'' \emph{arXiv preprint arXiv:1807.03748},
  2018.

\bibitem{ye2019unsupervised}
M.~Ye, X.~Zhang, P.~C. Yuen, and S.-F. Chang, ``Unsupervised embedding learning
  via invariant and spreading instance feature,'' in \emph{CVPR}, 2019.

\bibitem{tian2019contrastive}
Y.~Tian, D.~Krishnan, and P.~Isola, ``Contrastive multiview coding,'' in
  \emph{ECCV}, 2019.

\bibitem{Wang_2021_CVPR}
X.~Wang, R.~Zhang, C.~Shen, T.~Kong, and L.~Li, ``Dense contrastive learning
  for self-supervised visual pre-training,'' in \emph{CVPR}, 2021.

\bibitem{ibot}
J.~Zhou, C.~Wei, H.~Wang, W.~Shen, C.~Xie, A.~Yuille, and T.~Kong, ``ibot:
  Image bert pre-training with online tokenizer,'' in \emph{ICLR}, 2022.

\bibitem{data2vec}
A.~Baevski, W.-N. Hsu, Q.~Xu, A.~Babu, J.~Gu, and M.~Auli, ``Data2vec: A
  general framework for self-supervised learning in speech, vision and
  language,'' \emph{arXiv preprint arXiv:2202.03555}, 2022.

\bibitem{wmasked}
C.~Wei, H.~Fan, S.~Xie, C.-Y. Wu, A.~Yuille, and C.~Feichtenhofer, ``Masked
  feature prediction for self-supervised visual pre-training,'' in \emph{CVPR},
  2022.

\bibitem{t5}
C.~Raffel, N.~Shazeer, A.~Roberts, K.~Lee, S.~Narang, M.~Matena, Y.~Zhou,
  W.~Li, and P.~J. Liu, ``Exploring the limits of transfer learning with a
  unified text-to-text transformer,'' \emph{Journal of Machine Learning
  Research}, 2020.

\bibitem{gpt}
T.~Brown, B.~Mann, N.~Ryder, M.~Subbiah, J.~D. Kaplan, P.~Dhariwal,
  A.~Neelakantan, P.~Shyam, G.~Sastry, A.~Askell \emph{et~al.}, ``Language
  models are few-shot learners,'' in \emph{NeurIPS}, 2020.

\bibitem{he2016deep}
K.~He, X.~Zhang, S.~Ren, and J.~Sun, ``Deep residual learning for image
  recognition,'' in \emph{CVPR}, 2016.

\bibitem{glip}
L.~H. Li, P.~Zhang, H.~Zhang, J.~Yang, C.~Li, Y.~Zhong, L.~Wang, L.~Yuan,
  L.~Zhang, J.-N. Hwang \emph{et~al.}, ``Grounded language-image
  pre-training,'' in \emph{CVPR}, 2022.

\bibitem{loctex}
Z.~Liu, S.~Stent, J.~Li, J.~Gideon, and S.~Han, ``Loctex: Learning
  data-efficient visual representations from localized textual supervision,''
  in \emph{ICCV}, 2021.

\bibitem{girshick2015fast}
R.~Girshick, ``Fast r-cnn,'' in \emph{ICCV}, 2015.

\bibitem{position}
A.~J. Wang, P.~Zhou, M.~Z. Shou, and S.~Yan, ``Position-guided text prompt for
  vision-language pre-training,'' in \emph{CVPR}, 2023.

\bibitem{itm}
A.~Radford, J.~W. Kim, C.~Hallacy, A.~Ramesh, G.~Goh, S.~Agarwal, G.~Sastry,
  A.~Askell, P.~Mishkin, J.~Clark \emph{et~al.}, ``Learning transferable visual
  models from natural language supervision,'' in \emph{ICML}, 2021.

\bibitem{sharma2018conceptual}
P.~Sharma, N.~Ding, S.~Goodman, and R.~Soricut, ``Conceptual captions: A
  cleaned, hypernymed, image alt-text dataset for automatic image captioning,''
  in \emph{ACL}, 2018.

\bibitem{He_2016_CVPR}
K.~He, X.~Zhang, S.~Ren, and J.~Sun, ``Deep residual learning for image
  recognition,'' in \emph{CVPR}, 2016.

\bibitem{He_2017_ICCV}
K.~He, G.~Gkioxari, P.~Dollar, and R.~Girshick, ``Mask r-cnn,'' in \emph{ICCV},
  2017.

\bibitem{esser2021taming}
P.~Esser, R.~Rombach, and B.~Ommer, ``Taming transformers for high-resolution
  image synthesis,'' in \emph{CVPR}, 2021.

\bibitem{ramesh2021zero}
A.~Ramesh, M.~Pavlov, G.~Goh, S.~Gray, C.~Voss, A.~Radford, M.~Chen, and
  I.~Sutskever, ``Zero-shot text-to-image generation,'' in \emph{ICML}, 2021.

\bibitem{long2015fully}
J.~Long, E.~Shelhamer, and T.~Darrell, ``Fully convolutional networks for
  semantic segmentation,'' in \emph{CVPR}, 2015.

\bibitem{vit}
A.~Dosovitskiy, L.~Beyer, A.~Kolesnikov, D.~Weissenborn, X.~Zhai,
  T.~Unterthiner, M.~Dehghani, M.~Minderer, G.~Heigold, S.~Gelly \emph{et~al.},
  ``An image is worth 16x16 words: Transformers for image recognition at
  scale,'' in \emph{ICLR}, 2021.

\bibitem{uper}
T.~Xiao, Y.~Liu, B.~Zhou, Y.~Jiang, and J.~Sun, ``Unified perceptual parsing
  for scene understanding,'' in \emph{ECCV}, 2018.

\end{thebibliography}
}

%%%%%%%%%%%%%%%%%%%%%%%%%%%%%%%%%%%%%%%%%%%%%%%%%%%%%%%%%%%%

\end{document}